\documentclass[11pt]{article}

\usepackage[preprint]{acl} 
\usepackage{times}
\usepackage{latexsym}
\usepackage[T1]{fontenc}
\usepackage[utf8]{inputenc}
\usepackage{microtype}
\usepackage{inconsolata}
\usepackage{graphicx}
\usepackage{booktabs}
\usepackage{multirow}
\usepackage{array}
\usepackage{amsmath}
\usepackage{amssymb}
\usepackage{xspace}
\usepackage{enumitem}
\usepackage{xcolor}
\usepackage{mathtools}
\usepackage{pdflscape}

\newcommand{\densemodel}{\textsc{mDenseOn}\xspace}
\newcommand{\lateon}{\textsc{mLateOn}\xspace}
\newcommand{\denseonen}{\textsc{DenseOn}\xspace}
\newcommand{\lateonen}{\textsc{LateOn}\xspace}
\newcommand{\baseencoder}{mmBERT-base\xspace}
\newcommand{\ndcg}{nDCG@10\xspace}

\title{DenseOn with the LateOn: Fully Open Dense and Late-Interaction Models\\
for Multilingual, Long-Context, and Code Search}

\author{
 \textbf{Raphaël Sourty\textsuperscript{*1}},
 \textbf{Antoine Chaffin\textsuperscript{*1}},
 \textbf{Paulo Roberto Moura Junior\textsuperscript{*1}},
 \textbf{Amélie Chatelain\textsuperscript{1}},
\\
 \textsuperscript{1}LightOn
\\
\small{\textsuperscript{*}Equal contribution}
\\
 \small{
   \textbf{Correspondence:} \href{mailto:raphael.sourty@lighton.ai}{raphael.sourty@lighton.ai},\href{mailto:antoine.chaffin@lighton.ai}{antoine.chaffin@lighton.ai},\href{mailto:paulo.demoura@lighton.ai}{paulo.demoura@lighton.ai},\href{mailto:amelie@lighton.ai}{amelie@lighton.ai}
 }
}

\begin{document}
\maketitle

\begin{abstract}
State-of-the-art retrieval models increasingly rely on closed training data, creating a reproducibility gap. We present an open end-to-end recipe for training retrieval models and study how English supervision transfers to multilingual retrieval through translate-train.
We first reconstruct and curate 665M English contrastive pre-training pairs from 1.4B pairs across 34 public sources and build 1.88M supervised fine-tuning pairs with mined hard negatives. Training yields two 149M-parameter models: \denseonen, a single-vector dense model, and \lateonen, a ColBERT-style late-interaction model. They achieve 56.20 and 57.22 average \ndcg on BEIR, respectively, setting new state-of-the-art results for this size class.
We then translate the validated English data into eight languages, yielding 2.8B pairs with cross-lingual samples, and train \densemodel and \lateon, two 307M-parameter models built on \baseencoder.
Despite sharing their backbone, data, and objectives, their representations behave differently: the dense model is strong on English and translated languages but degrades outside translate-train support, whereas the late-interaction model generalizes better to unseen languages and scripts. This suggests that token-level matching turns translate-train from a target-language expansion strategy into a multilingual generalization recipe. We publicly release the models, datasets, and training code.

\end{abstract}

\section{Introduction}

Modern retrieval models are no longer judged only on short English passage search: they are expected to work across languages, long documents, and specialized domains such as code. Recent dense and hybrid retrievers, including
gte-multilingual-base~\cite{zhang2024mgte},
BGE-M3~\cite{chen-etal-2024-m3},
Arctic-Embed 2.0~\cite{yu2024arcticembed2},
Qwen3-Embedding~\cite{zhang2025qwen3embeddingadvancingtext},
Jina v5 text~\cite{DBLP:conf/sigir/AkramSHHGWX26},
EmbeddingGemma~\cite{vera2025embeddinggemmapowerfullightweighttext},
Voyage 4~\cite{voyageai2026voyage4}, Harrier~\cite{huang2026harrier},
Granite Multilingual R2~\cite{awasthy2026graniteembeddingmultilingualr2},
and pplx-embed~\cite{eslami-etal-2026-diffusion},
show rapid progress across these settings. 
Late-interaction models such as Jina ColBERT v2~\cite{xiao-etal-2024-jina}, LFM2.5 ColBERT ~\cite{liquidAI2026Retrievers} and pplx-embed late~\cite{perplexity2026pplxlate} offer competitive multi-vector alternatives.
At the same time, their performance increasingly depends on large training mixtures, filtering pipelines, hard-negative mining, and teacher supervision that are only partially described or not released. As a result, it is often difficult to determine whether gains come from backbone architecture, scale, supervision, data quality, or benchmark overlap.

This is especially problematic because retrieval quality is often driven by the data recipe as much as other factors like the base model. A useful illustration is the gap between two models: modernbert-embed-base trained on the open Nomic Embed dataset~\cite{DBLP:journals/tmlr/NussbaumMMD25,nomicai2025modernbertembed}, and GTE-ModernBERT, trained on the closed GTE data mixture~\cite{zhang2024mgte,alibabanlp2025gtemodernbert}. Both use the ModernBERT backbone~\cite{warner-etal-2025-smarter}, yet the latter reaches almost 2.5 points higher average nDCG@10 on BEIR benchmark~\cite{beir_neurips}, as shown in Table~\ref{tab:decon-beir-rankings}. Since the backbone and training methodology are approximately held constant, the remaining performance gap most likely reflects differences in the training data and recipe. Moreover, open efforts such as Nomic Embed~\cite{DBLP:journals/tmlr/NussbaumMMD25} have shown the value of releasing retrieval data, enabling controlled follow-up work, for example, ColBERT-Zero~\cite{colbert-zero} compares dense models against late-interaction models under matched conditions. However, open recipes still lag behind the newest frontier systems, especially when multilingual, long-context, and code retrieval are considered together.

This work aims to rebuild the retrieval frontier on open foundations. We first reconstruct and validate a strong English retrieval recipe, then extend it to multilingual retrieval through translate-train, motivated by two observations: high-quality retrieval supervision remains concentrated in English~\cite{joshi-etal-2020-state}, and machine translation can transfer it effectively to other languages~\cite{DBLP:xnli, DBLP:pawsx, bonifacio2022mmarcomultilingualversionms}. In retrieval, translate-train has been shown to match full document translation at lower cost~\cite{DBLP:conf/ecir/LawrieYOM23}, refined via cross-encoder distillation~\cite{DBLP:conf/sigir/YangLM24}, and extended to low-resource languages~\cite{DBLP:journals/corr/abs-2404-08134}. However, prior work evaluates almost exclusively on the languages into which data was translated, leaving open whether translate-train generalizes \emph{beyond} its targets. We scale the strategy to a 2.8B-pair multilingual corpus and evaluate it across all 18 MIRACL languages, 13 of which are absent from training. We find that cross-lingual generalization depends strongly on the retrieval paradigm: late interaction transfers far more effectively to unseen languages and scripts than single-vector dense retrieval.

\paragraph{Contributions.}
\begin{itemize}[leftmargin=*]
    \item \textbf{An open English retrieval recipe at the frontier.}
    We reconstruct and curate a large-scale English retrieval training mixture from 34 public sources, releasing both the annotated full corpus and the curated training subset, together with a 1.88M-pair fine-tuning dataset with mined hard negatives. Every filtering step is non-destructive, meaning that anyone can build their own filters to add or replace ours without re-collecting the data. 

    \item \textbf{State-of-the-art open English retrievers at 149M parameters.}
    We train the 149M-parameter ModernBERT-base~\cite{warner-etal-2025-smarter} backbone with English data to create \denseonen and \lateonen, which achieve 56.20 and 57.22 average \ndcg on BEIR respectively, establishing new state-of-the-art results at their size class and validating the recipe before multilingual transfer.

    \item \textbf{A large-scale open translate-train pipeline for multilingual retrieval.}
    We extend the validated English recipe by translating the training data into eight target languages producing a 2.8B-pair multilingual pre-training corpus and a large multilingual fine-tuning dataset spanning nine languages. This also natively allows the creation of a very large quantity of cross-lingual paired data. We further extend the fine-tuning dataset with long-context and code data.

    \item \textbf{State-of-the-art open multilingual dense and late-interaction models.}
    We train paired dense and late-interaction models with multilingual data using the same 307M-parameter mmBERT-base~\cite{DBLP:journals/corr/abs-2509-06888} backbone. This controlled comparison shows that dense translate-train is strong mainly inside the translated-language support, whereas late interaction generalizes much more effectively to languages and scripts unseen during training. The resulting \densemodel and \lateon models achieve strong results across BEIR, MIRACL, MLDR, and MTEB Code.
\end{itemize}

All models, datasets, and training scripts are publicly released, and we provide ablations covering data mixture, KL distillation, Matryoshka training, and translate-train design choices.

\section{Method}
Our pipeline has two stages: an English stage that reconstructs and validates a strong retrieval data recipe, and a multilingual stage that transfers it through translate-train. This section describes the data construction and training procedures for both.

\begin{enumerate}[leftmargin=*]
    \item \textbf{English retrieval data and models.}
    We reassemble the English retrieval mixture described in the mGTE technical
    report~\cite{zhang2024mgte}, track down the 34 public sources, and apply a
    non-destructive multi-stage filtering pipeline, as detailed in
    Section~\ref{sec:english-pretraining}. We complement this pre-training corpus with a supervised fine-tuning dataset comprising hard negatives mined following NV-Retriever~\cite{moreira2024nvretriever} and annotated with cross-encoder scores (Section~\ref{sec:english-finetuning}). We train matched dense and late-interaction
    English models on the ModernBERT backbone~\cite{warner-etal-2025-smarter}.

    \item \textbf{Multilingual, long-context, and code retrieval.}
    We extend the English recipe by translating the curated data into eight target
    languages: French, German, Italian, Spanish, Portuguese, Swedish, Norwegian, and
    Arabic. We additionally construct cross-lingual pairs to support cross-lingual
    retrieval. Finally, we augment supervised fine-tuning data with organic multilingual, long-context, and
    code data from MIRACL~\cite{zhang-etal-2023-miracl}, MLDR~\cite{chen-etal-2024-m3}, and
    LateOn-Code~\cite{LateOn-Code}. Using the \baseencoder backbone
    ~\cite{DBLP:journals/corr/abs-2509-06888}, we train matched dense and late-interaction multilingual
    models.
\end{enumerate}

\subsection{English Data}
\label{sec:data}

\subsubsection{English Pre-Training Data}
\label{sec:english-pretraining}

\paragraph{Reconstructing the mGTE mixture.}

The mGTE technical report~\cite{zhang2024mgte} describes its data sources but does not release the data.
We reassemble the mixture from scratch, tracking down each of the 34
public sources, matching formats, and rebuilding the full pipeline.
To this, we add a new Common Crawl component derived from FineWeb-Edu~\cite{NEURIPS2024_370df50c}, which replaces the outdated web-crawl split used in the original report with more recent and higher quality data. We use the titles of the pages as queries and the content as documents.
The raw corpus contains approximately 1.4B query-document pairs before filtering.

\paragraph{Non-destructive filtering pipeline.}
Rather than discarding low-quality data, we annotate every pair with filtering signals
as metadata: a boolean structural-quality flag, an MD5-based deduplication column,
and a cross-encoder relevance score from mxbai-rerank-large-v2~\cite{DBLP:conf/acl/LiSHLCL26}.
Filtering is applied at training time, enabling downstream users to adjust thresholds or add their filters without rerunning the pipeline.

The structural filtering stage applies over 30 composable, source-aware filters, including surface-level clean-up (HTML, legal boilerplate, non-printable characters), language identification (FastText with $\geq 50\%$ confidence;~\citealp{joulin2017bag}), unicode-script consistency, statistical quality heuristics (Google Web 1T unigram log-probabilities from~\citealp{brants2006web1t}, repeated uncommon words, special-character ratios) and digit/uppercase ratio checks.
Per-source configuration is a key design decision: retention ranges from $\sim$49\%
on noisy news headlines to $\sim$97\% on curated scientific abstracts.

Semantic relevance filtering uses an absolute threshold of 3.0 on the
cross-encoder score: we experimented with various thresholds and ultimately selected one that enabled us to match the original mGTE mixture as closely as possible.
For FineWeb-Edu, which is already curated and comprises over half of the mixture, we skip
rule-based filtering and apply a more aggressive percentile threshold. This retains roughly the top third of pairs.

After filtering, the training-ready English corpus contains \textbf{665M pairs}
($\sim$48\% retention).
We release the full annotated corpus (all 1.4B pairs with all metadata), together with a curated version reflecting the training slice used in our experiments.

\subsubsection{English Fine-Tuning Data}
\label{sec:english-finetuning}

After large-scale contrastive pre-training, we refine the model on a smaller,
higher-quality dataset where each query is paired with a positive document and
specifically mined hard negatives.

We mine hard-negatives using the NV-Retriever~\cite{moreira2024nvretriever} approach across seven
widely-used datasets: FiQA~\cite{maia2018fiqa}, Natural Questions~\cite{kwiatkowski2019naturalquestions}, HotpotQA~\cite{yang2018hotpotqa}, MS MARCO~\cite{DBLP:conf/nips/NguyenRSGTMD16}, FEVER~\cite{thorne2018fever}, SQuADv2~\cite{rajpurkar2018squad2}, and TriviaQA~\cite{joshi2017triviaqa}.
For each query we retrieve the top-2,048 nearest passages with GTE-ModernBERT.
The resulting dataset contains \textbf{1.88M} pairs. We release it in full, including all 2,048 mined passages and their associated retrieval scores, enabling the community to apply their own filtering criteria.

\subsection{Translate-Train: Extending to Multilingual}
\label{sec:translate-train}
\subsubsection{Multilingual Pre-Training Data}

We extend the curated English data to eight target languages by machine translation, so that the multilingual corpus directly inherits the structural, deduplication, and relevance filtering of the English seed.

\paragraph{Translation pipeline.}
We translate each English query-document pair jointly into eight target languages: French, German, Italian, Spanish, Portuguese, Swedish, Norwegian, and Arabic. Joint translation keeps the query and document grounded in the same context, while language-specific prompts encourage idiomatic reformulation rather than literal transliteration. We use Mistral-Small-3.1-24B-Instruct~\cite{mistralai2025mistralsmall31} with two in-language examples per target language.

\paragraph{Cross-lingual pairs.}
To support cross-lingual retrieval, we convert a subset of monolingual translated pairs into cross-lingual pairs. For a query in language $\ell$, we replace the paired document with its translation into a uniformly sampled language $\ell' \neq \ell$, keeping the query unchanged. 

\paragraph{Resulting corpus.}
The multilingual pre-training corpus contains approximately \textbf{2.8B} query-document pairs, including a 25\% cross-lingual share per language (see Appendix~\ref{app:pt-data} for per-source details). We set the cross-lingual share of each multilingual split after ablating different proportions (Appendix~\ref{app:pretraining-mixture}). During pre-training, this multilingual corpus is mixed with the curated English seed corpus.

\subsubsection{Multilingual Fine-Tuning Data}

The multilingual fine-tuning corpus is built by translating the English fine-tuning
pairs into the eight target languages using 
Qwen3-32B~\cite{yang2025qwen3technicalreport}\footnote{This model was not available at the time of the pre-training translation}.  Before translation, we reduce the negative set by retaining exactly the top 10 mined hard negatives per English pair with NV-Retriever filtering of 0.95 (as in Section~\ref{sec:sft-setup}), thereby reducing the number of passages to translate.

As the scale of fine-tuning is much smaller than that of pre-training, we can afford to gather some data from outside our English seed. Notably, we augment our fine-tuning data with three sources:

\begin{enumerate}
    \item Multilingual data from MIRACL~\cite{zhang-etal-2023-miracl}, to leverage high-quality organic data and reduce the possible biases of using only machine-translated data;
    \item Long context data from MLDR~\cite{chen-etal-2024-m3}, which provides more organic data and increases the models' long-context capabilities, as most of the data we collected is relatively short context;
    \item Code data from LateOn-Code~\cite{LateOn-Code}, to give the model code retrieval capabilities. We use only the fine-tuning data because our pre-training experiments focus on the natural-language mixture. Adding CoRNStack~\cite{DBLP:conf/iclr/SureshRXNMDJ25} during pre-training may further improve code retrieval, and we leave this for future work.
\end{enumerate}

Although the additional multilingual datasets cover many languages, we restrict them to our target languages, leaving the remaining languages unseen during retrieval training.
For the remaining non-translated multilingual data, we follow the English mining procedure using arctic-embed-l-v2~\cite{yu2024arcticembed2}, retaining the top-2,048 candidates and their retriever scores. For code, the existing data already matches our format, with 2,048 hard negatives mined using GTE-ModernBERT. The existing code data covers all MTEB Code~\cite{ICLR2025_fc0e3f90} splits besides CodeEditSearch that lacks a training set. We build the training dataset from CommitPackFT~\cite{ICLR2024_1ec299a5} and decontaminate it against the CodeEditSearch evaluation set on MTEB by removing shared commit SHAs, exact normalized-text matches, and examples with at least 50\% 13-gram containment overlap~\cite{gpt3}.
We further annotate the pairs with mxbai-rerank-large-v2~\cite{DBLP:conf/acl/LiSHLCL26} to allow for distillation from cross-encoder scores.
The full dataset contains \textbf{16.3M} contrastive samples across nine natural languages as well as code. The distribution of samples for each language and data source is better detailed in Appendix~\ref{app:sft-data}.

\subsection{Training Setup}
For all training phases, including contrastive pre-training and supervised fine-tuning, we use Sentence Transformers~\cite{reimers-gurevych-2019-sentence} for dense models and PyLate~\cite{DBLP:conf/cikm/ChaffinS25} for late-interaction models. For dense models, we use [CLS] pooling, which we find to be stronger than mean pooling, and prepend asymmetric prefixes to queries and documents (\texttt{"query:"} and \texttt{"document:"}). Together, these two components yield a significant performance boost of about 1.5 average \ndcg on BEIR. As in the original ColBERT~\cite{DBLP:conf/sigir/KhattabZ20}, we also apply (shorter) asymmetric prefixes \texttt{"[Q]"} and \texttt{"[D]"}. Apart from these details and the retrieval paradigm, the models share the same backbone, training data, and objectives, enabling a controlled comparison.

\subsubsection{Contrastive Pre-Training Setup}

Pre-training uses multinomial language sampling~\cite{DBLP:conf/nips/ConneauL19, DBLP:conf/acl/ConneauKGCWGGOZ20, zhang2024mgte} with $\alpha=0.5$ to prevent large sources from dominating training. At each step, the batch is drawn from a single source so the model cannot exploit source-specific shortcuts~\cite{DBLP:journals/tmlr/NussbaumMMD25}. The mixture used is composed of 20\% English-only data and 80\% of multilingual data, with 25\%  cross-lingual pairs within multilingual split and rest are monolingual. We experiment with various mixtures for both the proportion of English over the other languages and the proportion of cross-lingual data within the multilingual split (Appendix~\ref{app:pretraining-mixture}). Interestingly, we observe little difference in performance. We also find that a large amount of English data is not necessary for good English performance, and that cross-lingual data do not necessarily improve multilingual performance. This suggests that a limited amount of cross-lingual data is sufficient, possibly because translate-train provides a strong alignment signal that leverages the multilingual alignment learned during mmBERT pretraining. 
We train with the contrastive InfoNCE loss~\cite{oord2019representationlearningcontrastivepredictive} using in-batch negatives. Since training uses only query-positive pairs, we rely on large batch sizes of 16k to construct informative batches, and use GradCache~\cite{gradcache} to keep memory requirements manageable. All the hyper-parameters can be found in Appendix~\ref{app:hyperparams}.

\subsubsection{Supervised Fine-Tuning Setup}
\label{sec:sft-setup}
After large-scale contrastive pre-training, we fine-tune the models on smaller, high-quality datasets. For each sample, we retain 10 negatives after filtering with NV-Retriever at a threshold of 0.95, from which we sample 7 random negatives at each step. Pairs with fewer negatives than required (after filtering) are discarded as likely weakly-matched. Both single-vector and multi-vector models are trained with mined hard negatives together with in-batch negatives. Since the negatives are explicitly mined, we use a smaller batch size than in pre-training, set to 128 samples. For single-vector fine-tuning, we apply Matryoshka training~\cite{DBLP:conf/nips/KusupatiBRWSRHC22} in the contrastive loss to support reduced embedding dimensions at inference time. Performance across truncated dimensions in Appendix~\ref{app:matryoshka}. For multi-vector fine-tuning, we use \textit{MeanMaxSim} as the scoring function, i.e., we normalize the MaxSim score by the query length. This slightly improves performance in our experiments (Appendix~\ref{app:meanmaxsim}) and helps reduce length biases, especially for longer queries.

We combine the contrastive objective with Knowledge Distillation (KD) to further improve model performance.
Following prior work~\cite{ren-etal-2021-rocketqav2, santhanam-etal-2022-colbertv2, xiao-etal-2024-jina, takehi2025mxbaiedgecolbert}, we distill a strong cross-encoder into our models by applying a KL divergence loss between student and teacher scores. We apply separate softmax temperatures for each distribution, as investigated in Appendix~\ref{app:distillation}. We use mxbai-rerank-large-v2~\cite{DBLP:conf/acl/LiSHLCL26} as teacher for both models. All the hyper-parameters can be found in Appendix~\ref{app:hyperparams}.

\section{Results}
\label{sec:results}
All the evaluations are run with the MTEB~\cite{muennighoff-etal-2023-mteb, ICLR2025_fc0e3f90} library to ensure correct evaluation and we use FastPlaid~\cite{fastplaid2025} to perform efficient ColBERT-style retrieval over large multi-vector indexes. Evaluation hyper-parameters are described in Appendix~\ref{app:eval-params}.

\begin{table}[t]
\centering
\setlength{\tabcolsep}{2pt}
\small
\begin{tabular}{lcccc}
\toprule
 & \multicolumn{2}{c}{Standard} & \multicolumn{2}{c}{Decont.} \\
\cmidrule(lr){2-3} \cmidrule(lr){4-5}
Model & BEIR & Rank & BEIR$_{14}$ & Rank \\
\midrule
\textbf{\lateonen}\dag      & \textbf{57.22} & 1  & \textbf{61.36} & 1~\textcolor{gray}{(--)} \\
pplx-embed-v1-0.6b          & 56.70          & 2  & 59.67          & 4~\textcolor{red}{($\downarrow$2)} \\
\textbf{\denseonen}         & \textbf{56.20} & 3  & \textbf{58.81} & 7~\textcolor{red}{($\downarrow$4)} \\
pplx-embed-v1-late-0.6b\dag & 56.11          & 4  & 59.84          & 3~\textcolor{teal}{($\uparrow$1)} \\
jina-v5-text-nano           & 56.08          & 5  & 58.83          & 6~\textcolor{red}{($\downarrow$1)} \\
ColBERT-Zero\dag            & 55.82          & 6  & 59.95          & 2~\textcolor{teal}{($\uparrow$4)} \\
arctic-embed-l-v2           & 55.55          & 7  & 57.92          & 10~\textcolor{red}{($\downarrow$3)} \\
Qwen3-Embedding-0.6B        & 55.33          & 8  & 56.96          & 11~\textcolor{red}{($\downarrow$3)} \\
GTE-ModernBERT              & 55.19          & 9  & 56.63          & 12~\textcolor{red}{($\downarrow$3)} \\
harrier-oss-v1-0.6b         & 55.04          & 10 & 57.98          & 9~\textcolor{teal}{($\uparrow$1)} \\
GTE-ModernColBERT\dag       & 54.23          & 11 & 59.29          & 5~\textcolor{teal}{($\uparrow$6)} \\
colbert-small\dag           & 53.79          & 12 & 58.05          & 8~\textcolor{teal}{($\uparrow$4)} \\
modernbert-embed-base       & 52.89          & 13 & 55.61          & 13~\textcolor{gray}{(--)} \\
\bottomrule
\end{tabular}
\caption{Ranking comparison between standard BEIR and decontaminated BEIR. The
final column reports the rank change after decontamination. Bold indicates our
models; \dag\ marks multi-vector models. We exclude CQADupstack from the decontaminated analysis because the overlap with training
data is too big and too few samples remained after filtering.}
\label{tab:decon-beir-rankings}
\end{table}

\begin{table*}[th]
\centering
\setlength{\tabcolsep}{4pt}
\small
\begin{tabular}{lccccccc}
\toprule
\textbf{Model}
  & \textbf{Size}
  & \textbf{BEIR}
  & \textbf{MIRACL\textsubscript{tgt}}
  & \textbf{MIRACL}
  & \textbf{MLDR\textsubscript{tgt}}
  & \textbf{MLDR}
  & \textbf{Code} \\
\midrule
\multicolumn{8}{l}{\textit{Our models}} \\
\lateon   & 307M & \textbf{57.56} & \textbf{65.61} & 67.04 & \textbf{87.69} & \textbf{77.92} & 73.48 \\
\densemodel & 307M & 56.70 & 59.61 & 58.02 & 64.98 & 51.59 & 71.53 \\
\midrule
\multicolumn{8}{l}{\textit{Dense baselines}} \\
pplx-embed-v1-0.6b            & 596M & 56.70 & 63.18 & 68.20 & 57.08 & 43.98 & 75.18 \\
jina-v5-text-small             & 677M & 56.70 & 61.03 & 66.56 & 53.00 & 43.83 & 73.01 \\
jina-v5-text-nano              & 239M & 56.08 & 60.86 & 65.84 & 56.71 & 47.20 & 70.75 \\
arctic-embed-l-v2            & 568M & 55.55 & 60.78 & 66.53 & 57.01 & 48.03 & 53.17 \\
Qwen3-Embedding-0.6B           & 596M & 55.33 & 57.64 & 60.62 & 59.11 & 50.06 & 73.75 \\
harrier-oss-v1-0.6b            & 596M & 55.04 & 57.41 & 63.68 & 53.84 & 44.23 & 70.97 \\
harrier-oss-v1-270m            & 268M & 50.82 & 48.95 & 57.11 & 52.31 & 42.63 & 63.79 \\
embeddinggemma-300m            & 308M & 53.69 & 58.72 & 64.58 & 49.11 & 40.89 & 68.81 \\
gte-multilingual-base          & 305M & 51.08 & 57.39 & 64.14 & 66.41 & 56.65 & 57.46 \\
voyage-4-nano                  & 340M & 49.96 & 49.29 & 58.65 & 62.10 & 51.97 & \textbf{76.43} \\
BGE-M3                         & 568M & 48.78 & 62.22 & \textbf{69.62} & 61.16 & 52.47 & 51.49 \\
granite-311m-r2                & 312M & 49.24 & 53.38 & 59.74 & 50.53 & 41.86 & 63.50 \\
\midrule
\multicolumn{8}{l}{\textit{Late-interaction baselines}} \\
pplx-embed-v1-late-0.6b        & 596M & 56.11 & 63.67 & 69.55 & 68.69 & 55.51 & 63.29 \\
LFM2.5-ColBERT-350M             & 353M & 54.50 & 57.40 & 42.45 & 78.39 & 61.05 & 51.62 \\
jina-colbert-v2                & 559M & 52.96 & 62.19 & 65.65 & 13.97 & 11.54 & 49.88 \\
GTE-ModernColBERT             & 149M & 54.23 & 43.96 & 36.35 & 67.68 & 45.95 & 54.37 \\
ColBERT-Zero                  & 149M & 55.82 & 43.89 & 39.36 & 71.32 & 49.50 & 53.59 \\
\bottomrule
\end{tabular}
\caption{Average retrieval results (\ndcg) on BEIR, MIRACL, MLDR, and MTEB Code.
\textbf{MIRACL\textsubscript{tgt}} and \textbf{MLDR\textsubscript{tgt}} include only benchmark languages that overlap with our retrieval-training languages. \textbf{MIRACL} and \textbf{MLDR} include all benchmark languages. Best overall per dataset in \textbf{bold}.}
\label{tab:headline}
\end{table*}

\subsection{English Results}
\label{sec:english-results}

We select the English pre-training mixture through a full-pipeline search over filtering thresholds and mixture compositions. Since pre-training retrieval scores are not always reliable predictors of post-fine-tuning quality~\cite{DBLP:journals/corr/abs-2405-05374}, each candidate recipe is evaluated after supervised fine-tuning on the full BEIR suite.

Using the selected mixture, \denseonen achieves 56.20 average \ndcg on BEIR, outperforming other base-size dense retrievers and remaining competitive with substantially larger models. \lateonen achieves 57.22 average \ndcg, surpassing existing ColBERT-style baselines at this size and outperforming several larger late-interaction models.

Furthermore, decontamination experiments confirm that these strong results stem from generalization and not benchmark leakage. Indeed, releasing data allows auditability and exploration of potential biases. For example, we can compute the contamination of BEIR with respect to our training data using exact hash matching plus 13-gram containment ($\geq$50\% overlap) following~\citet{gpt3}. The corpora range from near-intact (ArguAna: 1.5\% removed) to heavily reduced (NQ: 88.6\%
removed due to its Wikipedia basis). All the results on BEIR and decontaminated BEIR are reported in Table~\ref{tab:decon-beir-rankings}, with full per-dataset results in Appendix~\ref{app:decon-beir-extended}. Since decontamination removes evaluation examples that overlap with our training data, the resulting benchmark can be easier and raw scores may increase. The main signal is therefore the relative ranking before and after decontamination, rather than the absolute score difference.

The clearest pattern separates dense from late-interaction retrieval: every late-interaction model holds or improves its relative ranking after decontamination, while nearly every dense model drops. \lateonen remains the top-ranked model overall, and ColBERT-Zero, GTE-ModernColBERT~\cite{GTE-ModernColBERT}, and colbert-small~\cite{clavie2024smallcolbert} all gain between 1 and 6 ranks. This is consistent with prior observations that multi-vector models generalize better beyond their training data~\cite{santhanam-etal-2022-colbertv2,clavie2025jacolbert}, and foreshadows the same pattern we observe in the multilingual setting (Section~\ref{sec:multilingual-results}). \denseonen drops from rank 3 to rank 7 overall, but this shift is entirely driven by late-interaction models leapfrogging it: among dense models, it moves from second to third (flipping with jina-v5-text-nano by only 0.02 points). This robustness is particularly notable because decontamination is computed with respect to \textit{our} training data, making this the most adversarial setup for our models. This confirms that the English data recipe is not reliant on benchmark overlap, validating the mixture before we extend it through translate-train.

\subsection{Multilingual Results}
\label{sec:multilingual-results}
We evaluate the multilingual models, \densemodel and \lateon, along four axes: English general-domain retrieval on BEIR~\cite{beir_neurips}, multilingual retrieval on MIRACL~\cite{zhang-etal-2023-miracl}, long-document multilingual retrieval on MLDR~\cite{chen-etal-2024-m3}, and code retrieval on MTEB Code~\cite{ICLR2025_fc0e3f90}. Results are summarized in Table~\ref{tab:headline}.

For MIRACL and MLDR, we report two averages. Our retrieval training covers nine natural languages: English plus eight translate-train target languages. These target languages are Arabic, French, German, Italian, Norwegian, Portuguese, Spanish, and Swedish. The target-language average (MIRACL\textsubscript{tgt}, MLDR\textsubscript{tgt}) includes only benchmark languages that overlap with this retrieval-training set. This corresponds to Arabic, English, French, German, and Spanish for MIRACL, and Arabic, English, French, German, Italian, Portuguese, and Spanish for MLDR. The \emph{full} average includes all benchmark languages. Therefore, the gap between target-language and full averages measures transfer to languages not used during our retrieval training.\footnote{These languages may still have been seen during masked language modeling pre-training of the backbone, but not during our retrieval training stages.}

\paragraph{English retrieval (BEIR).}
\densemodel reaches 56.70 average \ndcg on BEIR, slightly surpassing the English-only \denseonen (56.20). This shows that multilingual training can preserve English retrieval quality, consistent with prior work on multilingual embeddings~\cite{yu2024arcticembed2}. \densemodel matches substantially larger dense systems such as pplx-embed-v1-0.6b and jina-v5-text-small (both 56.70) and outperforms all dense baselines of comparable size, including embeddinggemma-300m, gte-multilingual-base, and voyage-4-nano. \lateon achieves 57.56 average \ndcg, the highest BEIR score among all the evaluated multilingual models in Table~\ref{tab:headline}. It also slightly improves over the English-only \lateonen at 57.22 on BEIR. Per-task results are provided in Appendix~\ref{app:beir}.

\paragraph{Multilingual retrieval (MIRACL).}

On the target-language average, \densemodel reaches 59.61, matching models of similar size such as embeddinggemma-300m and trailing larger systems such as pplx-embed-v1-0.6b and BGE-M3
(Table~\ref{tab:headline}). On the full MIRACL average, however, \densemodel drops to 58.02, which is expected because only five MIRACL languages overlap with our retrieval-training languages (Arabic, English, French, German, and Spanish).

\lateon performs substantially better in both settings. It reaches \textbf{65.61} on the target-language average, the best score in Table~\ref{tab:headline}, and 67.04 on the full MIRACL average (9 points better than \densemodel). This places it close to substantially larger systems such as BGE-M3 and pplx-embed-v1-late-0.6b, despite using fewer retrieval training languages and roughly half as many parameters. These results show that late-interaction transfers better than a single-vector representation to languages outside the translate-train set. Per-language results are provided in Appendix~\ref{app:miracl}.

\paragraph{Long-document retrieval (MLDR).}
Before discussing MLDR results, we note that our fine-tuning data includes the MLDR training split for the target languages, whereas some baselines do not specify whether MLDR data was used during training. Comparisons with these baselines are therefore not fully controlled. The most reliable comparison is between \densemodel and \lateon, which share the same backbone and training data but differ in retrieval paradigm.

With that in mind, MLDR amplifies the pattern observed on MIRACL. On the target-language average, \densemodel reaches 64.98, competitive with most dense baselines in Table~\ref{tab:headline}. On the full MLDR average, however, it drops to 51.59, again reflecting weaker transfer to benchmark languages outside our retrieval-training set.

\lateon performs substantially better. It achieves \textbf{87.69} on MLDR\textsubscript{tgt} and \textbf{77.92} on full MLDR, the best scores in Table~\ref{tab:headline}, outperforming gte-multilingual-base, Qwen3-Embedding-0.6B, BGE-M3, and both pplx variants, all of which have documented exposure to the MLDR training set. Notably, \densemodel benefits from the same in-domain data advantage yet only matches the strongest dense baselines, showing that MLDR exposure alone does not explain the gap: with identical training data, late interaction outperforms dense retrieval by 20 \ndcg points. This advantage likely combines the stronger multilingual generalization of late-interaction with the well-documented affinity of ColBERT-style models for long-context retrieval~\cite{warner-etal-2025-smarter, GTE-ModernColBERT, takehi2025mxbaiedgecolbert}, where fine-grained token-level matching can exploit longer documents more effectively than a single pooled vector. Full per-language results are provided in Appendix~\ref{app:mldr}.

\paragraph{Code retrieval (MTEB Code).}
Despite using only fine-tuning data and no code-specific pre-training, both models achieve strong code retrieval performance. \densemodel scores 71.53 and \lateon 73.48, with \lateon behind only pplx-embed-v1-0.6b (75.18) and voyage-4-nano (76.43) and matching Qwen3-Embedding-0.6B (73.75).
\densemodel ranks second among dense models under 350M parameters, behind only voyage-4-nano. This demonstrates that targeted fine-tuning on code data can yield competitive performance even without a large-scale code pre-training corpus. Incorporating it at pre-training time may yield further gains. Per-task results are provided in Appendix~\ref{app:code}.

\paragraph{Late-interaction enables translate-train beyond target languages.}
Taken together, these results reveal a clear pattern.  Applied to a dense encoder, the translate-train recipe produces a model with strong
performance on in-distribution evaluation (English and its target languages), but with
limited generalization beyond them. Crucially, the dense model's degradation is not confined to unseen scripts: on MIRACL, \densemodel drops sharply on Latin-script languages absent from training such as Finnish, Indonesian, and Swahili, suggesting that the single-vector bottleneck struggles to transfer retrieval competence even when script-level overlap is high. \lateon, by contrast, generalizes to unseen Latin-script languages and to entirely different scripts (Cyrillic, CJK, Devanagari, Bengali, Telugu, Thai), achieving a higher score on full MIRACL (67.04) than on MIRACL\textsubscript{tgt} (65.61). However, it is important to note that the generalization is not uniform: languages that are likely underrepresented in the backbone pre-training, such as Indonesian and Swahili, remain weaker, indicating that late interaction broadens but does not eliminate the dependency on the base backbone language coverage. Enhancing the backbone's capabilities for these languages or adding a small amount of organic data (or at least some translated data) for these specific target languages would likely help close this gap~\cite{DBLP:conf/sigir/YangLM24}. Indeed, we restricted the data to a few target languages to study the generalization capabilities, although some data for these languages is available. Nevertheless, these results suggest that late interaction better preserves the multilingual structure learned during masked language modeling pre-training, turning translate-train from a target-language data expansion strategy into a practical path toward multilingual retrieval without exhaustively translating into every target language.

\section{Conclusion}

This work reduces the reproducibility gap between open and closed retrieval systems by releasing data, models, and training code for frontier retrieval development. We reconstruct and curate a large-scale English retrieval corpus of 665M query-document pairs, filtered from 1.4B pairs, together with a supervised fine-tuning dataset of 1.88M pairs with mined hard negatives. All filtering steps are non-destructive, allowing downstream users to audit, modify, and extend the recipe without re-collecting the data. Training on this English recipe yields \denseonen and \lateonen, establishing new state-of-the-art results for open models at their size class.\\
A central finding of our work is that strong retrieval performance can be achieved with publicly released data, provided that the data recipe is sufficiently broad, carefully filtered, and auditable. Decontamination experiments further show that the models' performance reflects genuine generalization rather than benchmark leakage.\\
We then show that this validated English recipe can be extended to multilingual retrieval through translate-train. By translating the curated English corpus into eight target languages and adding cross-lingual pairs, we construct a 2.8B-pair multilingual pre-training corpus, complemented by a 16.3M-sample multilingual fine-tuning dataset spanning nine natural languages and code. Models trained on this data achieve strong results across English retrieval, multilingual retrieval, long-document retrieval, and code search.\\
The main lesson, however, is that translate-train transfers differently under dense and late-interaction retrieval. \densemodel performs strongly on English and target-language benchmarks, but generalizes less reliably to languages outside the translated training set. \lateon consistently outperforms it and transfers much better to unseen languages and scripts, enabling strong multilingual retrieval without translating supervision into every language.

\section*{Limitations}

Our multilingual data are derived predominantly from translated English supervision. Machine translation may introduce artifacts and produce language unlike naturally occurring queries and documents. Jointly translating queries and documents may also make positive pairs artificially easy because both sides inherit similar lexical and stylistic choices. Although we add organic MIRACL and MLDR data during fine-tuning, most multilingual pre-training supervision remains synthetic. We do not conduct human evaluation of translation quality or per-language and per-domain error analyses.

Retrieval training covers English and eight translated languages. Performance elsewhere therefore depends partly on representations learned during \baseencoder pre-training and varies substantially across languages. Our results do not establish universal multilingual coverage. Whether the findings transfer to other backbones, scales, and deployment settings remains open.

\section*{Ethical Considerations}
Retrieval quality varies across languages, which can lead to unequal access to relevant information. Translate-train seeks to reduce this gap by transferring abundant English supervision to languages with less training data. However, the method relies on machine translation, whose quality may itself be lower for underrepresented languages. It may therefore be least effective for the languages that stand to benefit most. English-derived supervision may preserve English-centric assumptions and flatten culturally specific distinctions. Each intended language and domain should therefore be evaluated before deployment.
Translation, scoring, and training at this scale require substantial computation and have associated environmental costs. We release the resulting datasets, annotations, models, and training code to reduce the need to repeat these costly steps and lower the computational cost of reproducing our results or building on this work.

\section*{Acknowledgments}
We thank Xin Zhang, Zach Nussbaum, Tom Aarsen, Bo Wang, Eugene Yang, Benjamin Clavié, Nandan Thakur, Oskar Hallström and Iacopo Poli for their valuable contributions and feedback on the English mixture. We thank Orion Weller for building the FineWeb-derived Common Crawl split as well as for his feedback and help. We also thank Eugene Yang for his feedback on adapting our English study to multilinguality through translate-train.

This work was granted access to the HPC resources of IDRIS under GENCI allocations AS011016449, A0181016214, and A0171015706 (Jean Zay supercomputer). We also acknowledge the Barcelona Supercomputing Center (BSC-CNS) for providing access to MareNostrum 5 under EuroHPC AI Factory Fast Lane project EHPC-AIF-2025FL01-445. The English and code retrieval aspect of this project received funding from the BPI Scribe project. This project's multilingual extension is also supported by the OpenEuroLLM project, co-funded by the Digital Europe Programme under GA no. 101195233.

\bibliography{mybib}

\appendix

\section{Data}
\subsection{Multilingual Pre-Training Data}
\label{app:pt-data}
\begin{figure*}[t]
\centering
\includegraphics[width=\linewidth]{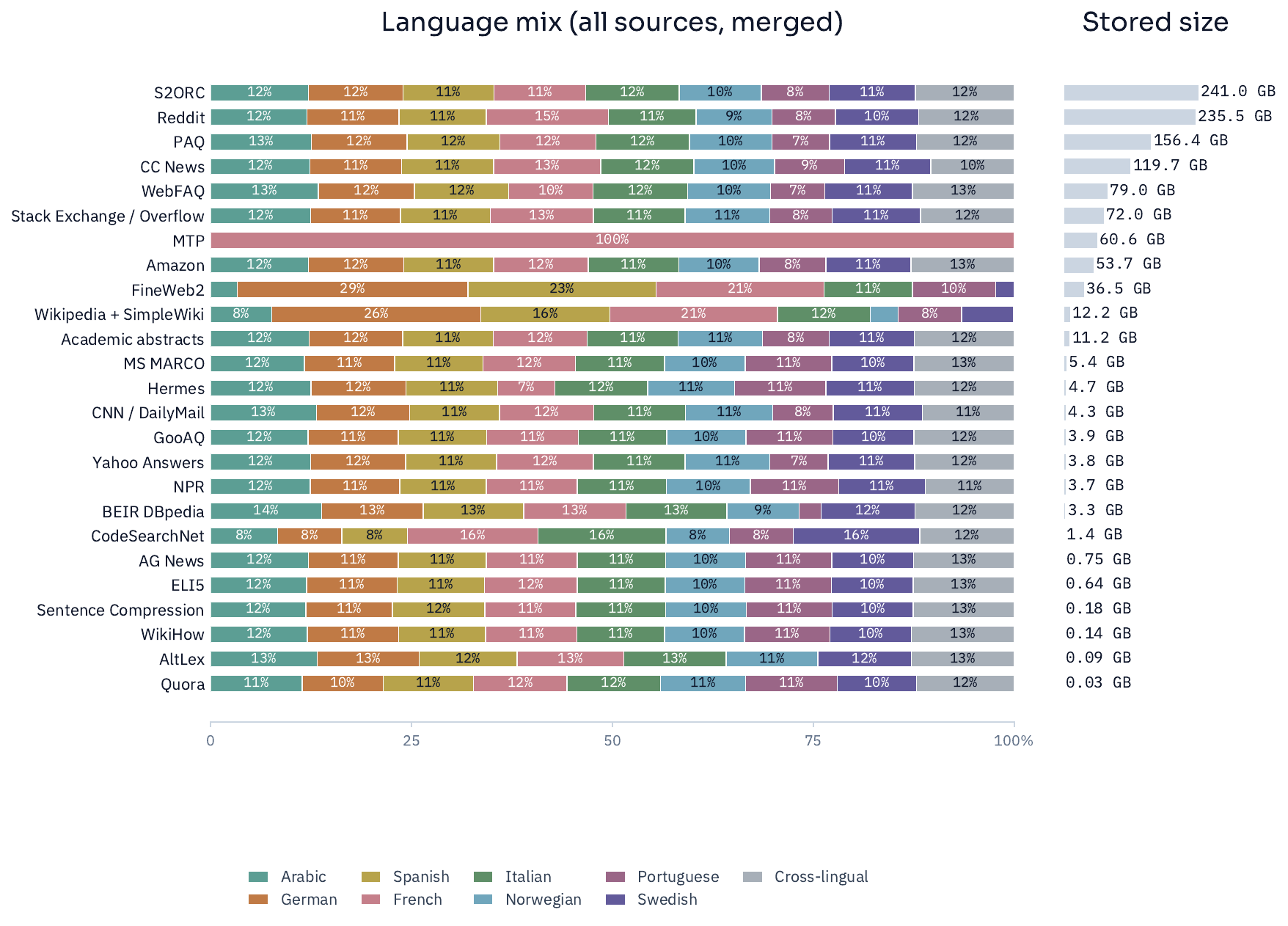}
\caption{Per-dataset language mix and stored size (GB) in the multilingual
contrastive pre-training data.}
\label{fig:pretraining-data}
\end{figure*}

Figure~\ref{fig:pretraining-data} details the language mix and stored size
(in GB) of each dataset in the multilingual contrastive pre-training corpus.

\subsection{Multilingual Fine-Tuning Data}
\label{app:sft-data}
\begin{figure*}[t]
\centering
\includegraphics[width=\linewidth]{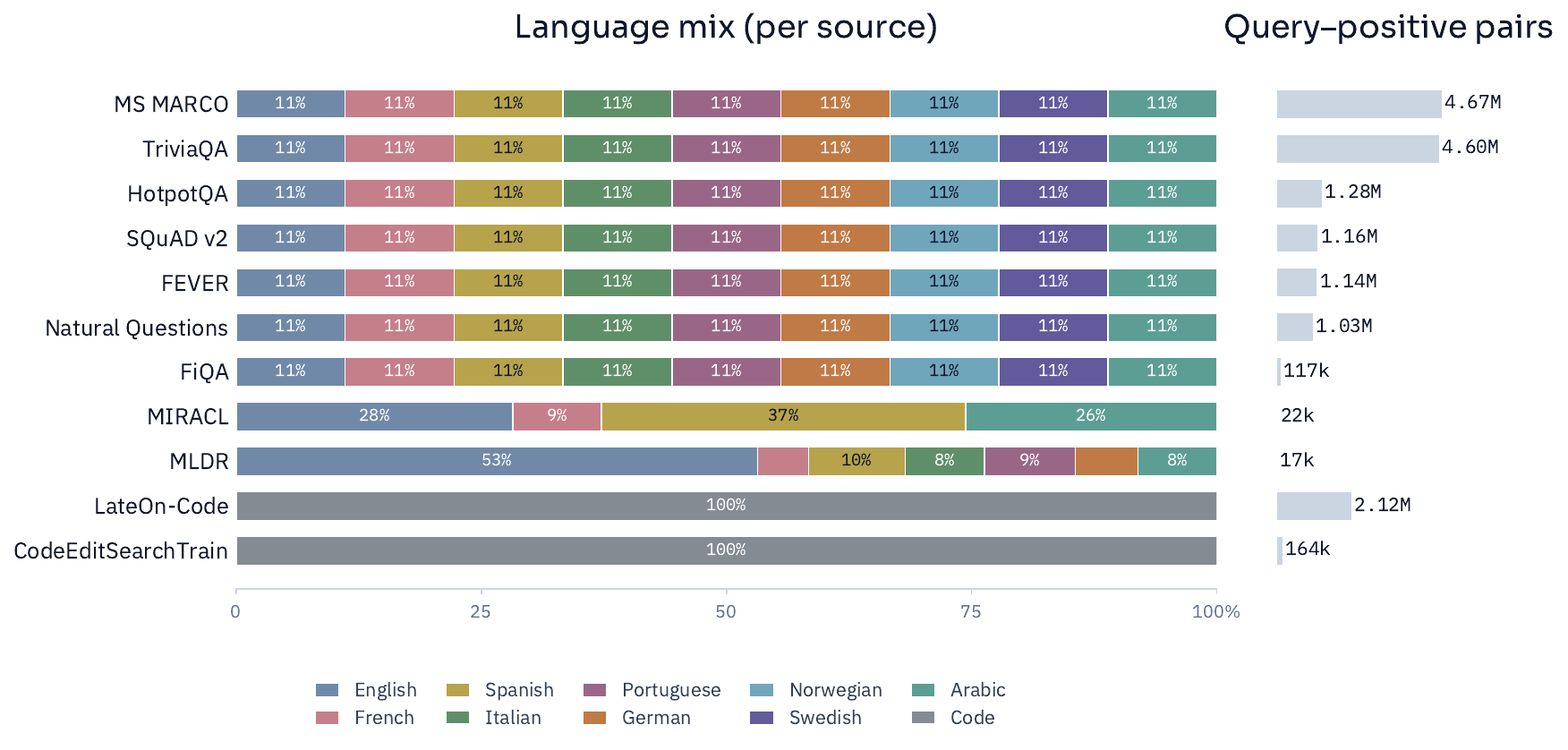}
\caption{Per-dataset language mix and total number of query-positive pairs in the filtered contrastive fine-tuning data.}
\label{fig:finetuning-data}
\end{figure*}

Figure~\ref{fig:finetuning-data} shows the language mix and the number of query-positive pairs of each dataset in the contrastive fine-tuning corpus, totaling 16.3M unique pairs.

\section{Detailed Results}
\label{app:extra-results}

\subsection{BEIR}
\label{app:beir}
Table~\ref{tab:beir-multilingual-full} report per-dataset BEIR results for all models.

\begin{table*}[t]
\centering
\footnotesize
\setlength{\tabcolsep}{2pt}
\begin{tabular}{lrrrrrrrrrrrrrrrr}
\toprule
\textbf{Model} & \textbf{Avg.} & AA & CQ & CF & DB & FE & FQ & HP & MS & NF & NQ & QU & SD & SF & TC & TO \\
\midrule
\multicolumn{17}{l}{\textit{Our models}} \\
\lateon     & \textbf{57.6} & 59.8 & 44.7 & 39.8 & \textbf{48.4} & \textbf{93.9} & 49.9 & \textbf{82.2} & 44.9 & 37.6 & \textbf{65.1} & 88.4 & 20.3 & 76.4 & 82.1 & 29.9 \\
\densemodel & 56.7 & 55.3 & 45.7 & 41.5 & 43.6 & 91.5 & 51.9 & 75.3 & 43.1 & 37.3 & 63.5 & 88.9 & 21.5 & 74.9 & 81.6 & 34.9 \\
\midrule
\multicolumn{17}{l}{\textit{Dense baselines}} \\
pplx-embed-v1-0.6b         & 56.7 & 60.5 & 46.0 & 39.8 & 44.3 & 90.7 & \textbf{52.1} & 74.4 & 43.9 & 35.8 & 62.0 & 89.0 & 22.8 & 74.8 & 85.6 & 29.0 \\
jina-v5-text-small   & 56.7 & 64.8 & 46.7 & 41.5 & 44.6 & 90.0 & 49.5 & 69.7 & 42.1 & \textbf{39.7} & 64.2 & \textbf{89.1} & 23.1 & 76.6 & 78.7 & 30.3 \\
jina-v5-text-nano    & 56.1 & 65.6 & 44.7 & 39.8 & 45.3 & 89.5 & 47.9 & 69.1 & 41.6 & 38.7 & 63.5 & 88.9 & 22.6 & 75.9 & 77.7 & 30.5 \\
arctic-embed-l-v2    & 55.6 & 58.8 & \textbf{47.2} & 38.1 & 43.9 & 91.7 & 44.2 & 72.4 & 44.0 & 35.9 & 64.7 & 89.1 & 20.3 & 72.3 & 80.3 & 30.4 \\
Qwen3-Embedding-0.6B & 55.3 & \textbf{69.3} & 45.8 & \textbf{42.1} & 39.5 & 88.0 & 47.4 & 65.7 & 38.1 & 36.2 & 53.4 & 88.1 & \textbf{24.3} & 69.9 & \textbf{89.5} & 33.0 \\
harrier-oss-v1-0.6b  & 55.0 & 66.0 & 44.9 & 25.7 & 45.6 & 80.2 & 47.7 & 72.3 & 42.7 & 38.3 & 60.2 & 88.5 & 23.2 & 76.6 & 82.6 & 31.3 \\
harrier-oss-v1-270m  & 50.8 & 63.0 & 40.8 & 22.7 & 42.5 & 73.6 & 39.4 & 68.1 & 40.0 & 36.3 & 56.5 & 86.7 & 21.3 & 72.9 & 70.0 & 28.4 \\
embeddinggemma-300m  & 53.7 & 66.0 & 42.1 & 26.7 & 44.5 & 81.1 & 47.4 & 70.1 & 38.6 & 39.2 & 63.4 & 86.6 & 19.5 & \textbf{78.7} & 76.7 & 24.8 \\
gte-multilingual-base & 51.1 & 58.2 & 38.1 & 34.8 & 40.1 & 92.1 & 45.1 & 63.0 & 40.0 & 36.7 & 58.1 & 88.1 & 18.2 & 73.5 & 57.4 & 22.8 \\
voyage-4-nano        & 50.0 & 58.7 & 44.3 & 22.2 & 39.5 & 68.4 & 50.9 & 62.0 & 31.6 & 39.6 & 49.1 & 86.1 & 21.4 & 75.2 & 77.8 & 22.5 \\
granite-311m-r2      & 49.2 & 57.2 & 43.2 & 29.9 & 34.4 & 82.2 & 39.9 & 62.3 & 31.5 & 31.0 & 55.8 & 87.3 & 21.9 & 71.0 & 68.3 & 22.8 \\
BGE-M3               & 48.8 & 54.0 & 39.1 & 29.6 & 39.8 & 81.5 & 41.3 & 69.4 & 38.3 & 31.4 & 60.6 & 88.6 & 16.3 & 64.4 & 54.9 & 22.6 \\
\midrule
\multicolumn{17}{l}{\textit{Late-interaction baselines}} \\
pplx-embed-v1-late-0.6b    & 56.1 & 55.2 & 46.7 & 35.4 & 46.7 & 92.6 & 51.4 & 80.3 & 45.0 & 37.5 & 61.7 & 87.0 & 20.6 & 73.9 & 80.3 & 27.4 \\
ColBERT-Zero         & 55.8 & 43.6 & 41.1 & 37.3 & 47.6 & 93.0 & 43.4 & 79.6 & 45.9 & 38.0 & 61.7 & 85.3 & 19.9 & 77.1 & 80.4 & \textbf{43.4} \\
LFM2.5-ColBERT-350M    & 54.5 & 40.9 & 41.1 & 40.5 & 42.5 & 92.2 & 49.5 & 80.6 & 45.8 & 37.0 & 60.5 & 86.1 & 18.2 & 74.5 & 80.2 & 27.9 \\
GTE-ModernColBERT & 54.2 & 37.8 & 36.9 & 33.1 & 47.7 & 91.5 & 43.4 & 78.0 & 45.8 & 37.6 & 61.6 & 86.7 & 18.4 & 75.0 & 85.2 & 34.6 \\
jina-colbert-v2      & 53.0 & 50.6 & 40.8 & 23.0 & 48.0 & 77.4 & 42.0 & 78.4 & \textbf{46.6} & 36.4 & 62.3 & 88.8 & 19.0 & 71.5 & 81.9 & 27.7 \\
\bottomrule
\end{tabular}
\caption{Per-dataset BEIR results (\ndcg) for our multilingual models and baselines. Best per column in \textbf{bold}. Dataset acronyms: AA (ArguAna), CQ (CQADupstack), CF (ClimateFEVER), DB (DBPedia), FE (FEVER), FQ (FiQA2018), HP (HotpotQA), MS (MSMARCO), NF (NFCorpus), NQ (Natural Questions), QU (Quora), SD (SCIDOCS), SF (SciFact), TC (TREC-COVID), TO (Touch\'e-2020).}
\label{tab:beir-multilingual-full}
\end{table*}

\subsection{Extended Decontaminated BEIR Analysis}

\label{app:decon-beir-extended}

Table~\ref{tab:decon-beir-extended} reports the full per-dataset decontaminated BEIR results used for the ranking comparison in Table~\ref{tab:decon-beir-rankings}. The decontaminated average is computed over
14 datasets: we exclude CQADupstack from the analysis because the overlap with training data is too big and too few samples remained after filtering. Despite the decontamination being adversarial to our models (as we use our training data to compute the overlap), \lateonen remains the strongest model globally and \denseonen among
the strongest dense retrievers. Overall, the comparison supports the main conclusion
that our models' gains are not explained by benchmark overlap. The ranking shift
also reinforces a pattern observed in the main paper: late-interaction models
tend to improve relative to single-vector models after decontamination.

\begin{table*}[t]
\centering
\footnotesize
\setlength{\tabcolsep}{1.75pt}
\begin{tabular}{lrrrrrrrrrrrrrrr}
\toprule
\textbf{Model} & \textbf{Avg.} & AA & CF & DB & FE & FQ & HP & MS & NF & NQ & QU & SD & SF & TC & TO \\
\midrule
\multicolumn{16}{l}{\textit{Our models}} \\
\lateonen$^\dagger$
& \textbf{61.4} & 52.2 & 42.1 & 31.7 & 92.7 & \textbf{57.9} & \textbf{78.9} & 70.3 & 27.0 & \textbf{93.1} & \textbf{91.5} & 15.1 & 88.9 & 80.9 & 36.8 \\
\denseonen
& 58.8 & 40.0 & 39.5 & 28.8 & 91.2 & 55.9 & 73.7 & 68.9 & 28.5 & 92.1 & 91.1 & 14.7 & 85.4 & 82.5 & 31.0 \\
\midrule
\multicolumn{16}{l}{\textit{Dense baselines}} \\
pplx-embed-v1-0.6b
& 59.7 & 43.7 & 42.4 & 28.4 & 91.1 & 55.2 & 73.5 & 71.9 & 28.0 & 91.6 & \textbf{91.5} & 15.4 & 89.0 & 83.7 & 30.0 \\
jina-v5-text-nano
& 58.8 & 47.2 & 41.6 & 30.2 & 90.0 & 51.5 & 67.5 & 68.6 & 29.4 & 92.3 & 91.3 & 14.9 & 89.4 & 76.8 & 33.2 \\
harrier-oss-v1-0.6b
& 58.0 & 47.4 & 25.7 & 31.3 & 80.7 & 50.1 & 71.4 & 73.4 & 27.9 & 90.0 & 90.9 & \textbf{17.1} & \textbf{90.7} & 81.8 & 33.3 \\
arctic-embed-l-v2
& 57.9 & 43.1 & 45.7 & 28.0 & 92.2 & 50.4 & 63.1 & 71.0 & 26.0 & 90.7 & 91.3 & 13.9 & 87.4 & 81.4 & 26.8 \\
bge-large-en-v1.5
& 57.3 & 46.0 & 39.0 & 28.9 & 87.6 & 49.3 & 75.2 & 68.9 & \textbf{29.8} & 85.9 & 91.3 & 14.0 & 86.5 & 72.7 & 26.9 \\
Qwen3-Embedding-0.6B
& 57.0 & 48.4 & 38.0 & 25.3 & 86.4 & 49.1 & 62.2 & 63.6 & 25.8 & 88.3 & 90.0 & 15.3 & 85.5 & \textbf{87.9} & 31.8 \\
GTE-ModernBERT
& 56.6 & 52.5 & \textbf{47.5} & 25.9 & \textbf{94.1} & 55.5 & 65.5 & 64.8 & 26.1 & 84.5 & 90.8 & 11.6 & 88.6 & 62.4 & 23.1 \\
bge-base-en-v1.5
& 56.2 & 45.6 & 32.9 & 26.7 & 86.8 & 44.5 & 72.7 & 66.8 & 27.4 & 85.6 & 91.1 & 13.8 & 87.6 & 76.6 & 28.1 \\
Nomic v1.5
& 55.9 & 35.8 & 43.5 & 28.8 & 86.8 & 44.7 & 72.7 & 67.4 & 24.4 & 85.1 & 87.2 & 12.7 & 83.3 & 80.7 & 29.4 \\
modernbert-embed-base
& 55.6 & 36.5 & 37.8 & 24.7 & 87.8 & 46.0 & 62.7 & 65.3 & 24.3 & 89.3 & 89.9 & 12.9 & 85.5 & 82.7 & 33.1 \\
\midrule
\multicolumn{16}{l}{\textit{Late-interaction baselines}} \\
ColBERT-Zero$^\dagger$
& 60.0 & 54.5 & 36.8 & 33.0 & 90.5 & 46.6 & 77.8 & \textbf{74.2} & 26.6 & 91.1 & 88.3 & 14.2 & 89.5 & 75.3 & \textbf{40.9} \\
pplx-embed-v1-late-0.6b$^\dagger$
& 59.8 & \textbf{60.9} & 36.4 & 29.9 & 89.7 & 50.9 & 78.6 & 69.2 & 27.9 & 92.8 & 83.8 & 13.5 & 89.3 & 80.2 & 34.7 \\
GTE-ModernColBERT$^\dagger$
& 59.3 & 48.8 & 33.5 & \textbf{33.2} & 88.1 & 50.2 & 77.3 & 71.6 & 27.3 & \textbf{93.1} & 89.1 & 13.6 & 87.7 & 81.4 & 35.3 \\
colbert-small$^\dagger$
& 58.1 & 47.7 & 35.7 & 31.7 & 89.3 & 45.6 & 77.1 & 71.4 & 25.0 & 86.2 & 90.1 & 13.1 & 89.2 & 81.5 & 29.0 \\
\bottomrule
\end{tabular}
\caption{Per-dataset decontaminated BEIR results (\ndcg). Our English models are grouped at the top; \textbf{Avg.} is computed over 14 datasets, used in the ranking comparison in Table~\ref{tab:decon-beir-rankings}.
$^\dagger$ marks multi-vector late-interaction models; best per column in \textbf{bold}. We exclude CQADupstack from the analysis because the overlap with training data is too large and too few samples remained after filtering. Dataset acronyms: AA (ArguAna), CF (ClimateFEVER), DB (DBPedia), FE (FEVER), FQ (FiQA2018), HP (HotpotQA), MS (MS MARCO), NF (NFCorpus), NQ (Natural Questions), QU (Quora), SD (SCIDOCS), SF (SciFact), TC (TREC-COVID), TO (Touch\'e-2020).}
\label{tab:decon-beir-extended}
\end{table*}

\subsection{MIRACL}
\label{app:miracl}

Table~\ref{tab:miracl-full} reports per-language MIRACL results for our multilingual models and other baselines. As noted in the main body, \densemodel is strong in-distribution (English and target languages), but drops sharply on Latin-script languages absent from training, such as Finnish (52.5), Indonesian (47.5), and Swahili (42.3), despite sharing the same script as the majority of its training data. \lateon not only outperforms \densemodel on in-distribution benchmarks but also generalizes substantially to unseen languages and scripts. Indeed, \lateon generalizes to unseen Latin-script languages such as Finnish~(74.3) and Yoruba~(69.1), and reaches strong performances on entirely different scripts, including Cyrillic~(Russian: 71.4), CJK~(Japanese: 72.1, Korean: 71.9, Chinese: 64.0), Devanagari~(Hindi: 60.4), Bengali~(73.4), Telugu~(71.8), and Thai~(76.0), all of which are unseen during retrieval training. Remarkably, \lateon achieves a higher score on full MIRACL (67.04) than on MIRACL\textsubscript{tgt} (65.61). However, it is important to note that the generalization is limited on languages that are likely underrepresented in the backbone pre-training, such as Indonesian~(55.4) and Swahili~(57.8) indicating that late interaction broadens but does not eliminate the dependency on the backbone's language coverage.

\begin{table*}[t]
\centering
\footnotesize
\setlength{\tabcolsep}{1.5pt}
\begin{tabular}{lrrrrrrrrrrrrrrrrrrr}
\toprule
\textbf{Model} & \textbf{Avg.} & ar & en & fr & es & de & bn & zh & fi & hi & id & ja & ko & fa & ru & sw & te & th & yo \\
\midrule
\multicolumn{20}{l}{\textit{Our models}} \\
\lateon     & 67.0 & 78.4 & \textbf{64.1} & \textbf{60.4} & \textbf{61.1} & \textbf{64.0} & 73.4 & 64.0 & 74.3 & 60.4 & 55.4 & 72.1 & 71.9 & 61.1 & 71.4 & 57.8 & 71.8 & 76.0 & 69.1 \\
\densemodel & 58.0 & 73.7 & 58.4 & 56.5 & 55.4 & 54.0 & 54.5 & 56.4 & 52.5 & 54.6 & 47.5 & 63.0 & 61.3 & 52.5 & 62.8 & 42.3 & 61.6 & 67.5 & 70.0 \\
\midrule
\multicolumn{20}{l}{\textit{Dense baselines}} \\
BGE-M3               & \textbf{69.6} & \textbf{78.9} & 57.8 & 59.5 & 57.3 & 57.7 & \textbf{79.9} & 63.7 & 77.8 & 60.1 & 56.1 & \textbf{73.1} & 70.2 & \textbf{61.3} & 70.3 & \textbf{78.6} & \textbf{86.3} & \textbf{82.7} & 82.0 \\
pplx-embed-v1-0.6b        & 68.2 & 78.0 & 57.4 & 59.8 & 60.0 & 60.7 & 75.2 & \textbf{65.3} & 76.7 & 61.5 & 54.9 & 70.8 & 70.3 & 60.0 & 69.5 & 74.1 & 72.6 & 78.7 & 82.1 \\
jina-v5-text-small   & 66.6 & 74.9 & 56.9 & 58.7 & 56.7 & 58.0 & 76.6 & 63.7 & 75.1 & 59.1 & 52.8 & 69.2 & 67.6 & 58.4 & 68.4 & 61.9 & 82.1 & 78.4 & 79.8 \\
arctic-embed-l-v2    & 66.5 & 76.7 & 54.3 & 57.4 & 56.5 & 59.0 & 74.0 & 61.3 & 76.6 & 59.1 & 52.4 & 66.5 & 66.2 & 60.7 & 67.5 & 70.8 & 82.6 & 77.5 & 78.4 \\
jina-v5-text-nano    & 65.8 & 73.8 & 56.1 & 58.4 & 56.6 & 59.4 & 73.1 & 62.5 & 74.1 & 59.3 & 52.1 & 69.0 & 65.8 & 58.8 & 66.5 & 63.3 & 77.4 & 77.4 & 81.6 \\
embeddinggemma-300m  & 64.6 & 69.3 & 57.1 & 56.3 & 55.2 & 55.7 & 74.5 & 63.5 & 69.3 & 58.4 & 51.0 & 65.9 & 64.8 & 57.7 & 64.5 & 70.5 & 79.5 & 74.8 & 74.4 \\
gte-multilingual-base & 64.1 & 71.8 & 54.8 & 55.6 & 53.4 & 51.2 & 72.6 & 63.4 & 73.5 & 52.4 & 50.5 & 66.5 & 63.9 & 54.9 & 63.9 & 69.7 & 83.1 & 74.5 & 79.0 \\
harrier-oss-v1-0.6b  & 63.7 & 74.0 & 52.5 & 53.5 & 54.6 & 52.4 & 71.6 & 56.5 & 72.2 & 58.7 & 50.8 & 63.5 & 64.6 & 56.0 & 61.1 & 67.7 & 79.8 & 74.9 & 81.9 \\
harrier-oss-v1-270m  & 57.1 & 69.2 & 46.1 & 39.4 & 49.2 & 40.9 & 66.8 & 47.3 & 66.7 & 54.4 & 45.7 & 54.3 & 55.2 & 52.1 & 52.1 & 63.9 & 76.5 & 69.5 & 78.8 \\
Qwen3-Embedding-0.6B & 60.6 & 70.2 & 52.1 & 55.8 & 55.7 & 54.4 & 66.3 & 58.2 & 70.5 & 52.0 & 51.6 & 63.2 & 60.3 & 52.9 & 61.6 & 45.4 & 76.4 & 73.6 & 71.1 \\
granite-311m-r2      & 59.7 & 67.9 & 45.2 & 51.9 & 50.9 & 50.9 & 70.8 & 58.3 & 68.5 & 51.8 & 46.9 & 62.6 & 58.8 & 52.4 & 56.8 & 66.2 & 80.2 & 71.8 & 63.4 \\
voyage-4-nano        & 58.7 & 65.1 & 45.2 & 42.5 & 47.4 & 46.3 & 70.2 & 51.3 & 69.1 & 55.5 & 46.5 & 57.7 & 61.7 & 52.4 & 55.2 & 66.1 & 78.7 & 71.8 & 73.1 \\
\midrule
\multicolumn{20}{l}{\textit{Late-interaction baselines}} \\
pplx-embed-v1-late-0.6b    & \textbf{69.6} & \textbf{78.9} & 59.7 & 57.8 & 60.9 & 61.0 & 76.4 & 61.8 & \textbf{78.1} & 64.0 & \textbf{57.4} & 72.7 & 70.2 & 60.1 & \textbf{72.9} & 75.7 & 76.9 & 79.9 & \textbf{87.4} \\
jina-colbert-v2      & 65.7 & 77.4 & 58.2 & 59.2 & 59.7 & 56.4 & 75.6 & 58.3 & 74.8 & \textbf{64.7} & 56.6 & 68.3 & \textbf{73.6} & 59.3 & 67.8 & 51.2 & 77.2 & 78.9 & 64.6 \\
LFM2.5-ColBERT-350M    & 42.5 & 71.1 & 58.4 & 51.5 & 53.0 & 53.0 & 9.7  & 45.1 & 46.1 & 20.2 & 33.3 & 64.3 & 68.2 & 29.1 & 55.1 & 41.4 & 0.3  & 10.9 & 53.4 \\
ColBERT-Zero         & 39.4 & 48.6 & 61.5 & 32.4 & 44.6 & 32.4 & 8.8  & 35.3 & 55.7 & 21.3 & 39.3 & 42.6 & 42.8 & 27.9 & 43.6 & 46.9 & 24.6 & 36.1 & 64.3 \\
GTE-ModernColBERT & 36.4 & 35.8 & 61.6 & 43.7 & 43.6 & 35.1 & 10.3 & 38.8 & 51.7 & 11.8 & 35.4 & 39.3 & 37.7 & 25.0 & 48.3 & 46.9 & 1.5  & 25.2 & 62.7 \\
\bottomrule
\end{tabular}
\caption{Per-language MIRACL results (\ndcg) for our multilingual models and baselines. Best per column in \textbf{bold}. Language codes: ar (Arabic), en (English), fr (French), es (Spanish), de (German), bn (Bengali), zh (Chinese), fi (Finnish), hi (Hindi), id (Indonesian), ja (Japanese), ko (Korean), fa (Persian), ru (Russian), sw (Swahili), te (Telugu), th (Thai), yo (Yoruba).}
\label{tab:miracl-full}
\end{table*}

\subsection{MLDR}
\label{app:mldr}

Table~\ref{tab:mldr-full} reports per-language long-context results on MLDR for our multilingual models and other baselines. On MLDR, the long-context setting amplifies the gap between single and multi-vector models:
\densemodel collapses on unseen languages such as Chinese~(18.8), Hindi~(27.2), and
Thai~(22.6), while \lateon remains functional across the board, reaching 54.8, 61.8,
and 51.4 on the same languages, thanks to the language generalization and long-context capabilities.

\begin{table*}[t]
\centering
\footnotesize
\setlength{\tabcolsep}{2pt}
\begin{tabular}{lrrrrrrrrrrrrrr}
\toprule
\textbf{Model} & \textbf{Avg.} & ar & en & fr & de & it & pt & es & zh & hi & ja & ko & ru & th \\
\midrule
\multicolumn{15}{l}{\textit{Our models}} \\
\lateon     & \textbf{77.9} & \textbf{86.8} & \textbf{80.8} & \textbf{94.2} & \textbf{70.7} & \textbf{90.9} & \textbf{92.0} & \textbf{98.5} & \textbf{54.8} & \textbf{61.8} & \textbf{73.5} & \textbf{68.4} & \textbf{89.2} & \textbf{51.4} \\
\densemodel & 51.6 & 50.7 & 54.2 & 78.1 & 46.6 & 66.1 & 77.8 & 81.5 & 18.8 & 27.2 & 50.4 & 35.3 & 61.5 & 22.6 \\
\midrule
\multicolumn{15}{l}{\textit{Dense baselines}} \\
gte-multilingual-base & 56.7 & 55.0 & 51.1 & 76.2 & 55.0 & 67.0 & 79.5 & 81.2 & 27.5 & 45.1 & 52.1 & 47.1 & 64.3 & 35.5 \\
BGE-M3               & 52.5 & 47.7 & 49.0 & 73.8 & 46.2 & 62.9 & 73.9 & 74.8 & 26.0 & 40.7 & 51.3 & 42.7 & 59.6 & 33.6 \\
voyage-4-nano        & 52.0 & 45.1 & 51.7 & 73.7 & 47.4 & 63.7 & 75.8 & 77.3 & 21.0 & 40.0 & 45.6 & 41.3 & 60.7 & 32.3 \\
Qwen3-Embedding-0.6B & 50.1 & 44.7 & 48.3 & 69.4 & 44.9 & 59.5 & 73.6 & 73.5 & 28.1 & 30.2 & 48.1 & 40.1 & 58.0 & 32.7 \\
arctic-embed-l-v2    & 48.0 & 40.6 & 38.6 & 67.9 & 45.4 & 60.8 & 72.2 & 73.7 & 21.1 & 31.5 & 47.7 & 36.9 & 56.9 & 31.2 \\
jina-v5-text-nano    & 47.2 & 42.2 & 43.7 & 68.7 & 40.3 & 58.8 & 71.5 & 71.7 & 19.1 & 30.5 & 48.0 & 36.5 & 54.3 & 28.3 \\
harrier-oss-v1-0.6b  & 44.2 & 37.6 & 46.9 & 67.7 & 31.5 & 57.6 & 67.4 & 68.2 & 20.6 & 27.0 & 42.7 & 31.7 & 53.0 & 23.2 \\
harrier-oss-v1-270m  & 42.6 & 35.3 & 44.5 & 65.8 & 30.3 & 56.8 & 65.4 & 68.0 & 17.0 & 30.2 & 39.1 & 32.9 & 49.7 & 19.0 \\
pplx-embed-v1-0.6b        & 44.0 & 39.7 & 45.2 & 71.8 & 43.0 & 57.9 & 69.6 & 72.5 & 20.7 & 2.2  & 42.5 & 32.7 & 51.5 & 22.6 \\
jina-v5-text-small   & 43.8 & 37.3 & 42.4 & 64.3 & 38.8 & 55.1 & 67.5 & 65.7 & 20.6 & 28.4 & 41.7 & 31.6 & 51.7 & 24.9 \\
granite-311m-r2      & 41.9 & 34.6 & 39.9 & 63.9 & 33.9 & 53.0 & 66.3 & 62.2 & 21.1 & 27.0 & 40.3 & 31.8 & 49.0 & 21.4 \\
embeddinggemma-300m  & 40.9 & 33.0 & 36.7 & 62.4 & 38.1 & 50.6 & 63.2 & 59.7 & 17.6 & 26.2 & 37.9 & 30.8 & 49.1 & 26.2 \\
\midrule
\multicolumn{15}{l}{\textit{Late-interaction baselines}} \\
LFM2.5-ColBERT-350M    & 61.1 & 66.2 & 67.4 & 89.0 & 64.7 & 80.9 & 87.8 & 92.9 & 31.8 & 2.2  & 62.0 & 59.8 & 79.0 & 10.1 \\
pplx-embed-v1-late-0.6b    & 55.5 & 54.9 & 59.7 & 80.2 & 51.5 & 68.7 & 80.5 & 85.4 & 36.5 & 13.7 & 56.1 & 41.8 & 63.9 & 28.8 \\
ColBERT-Zero         & 49.5 & 16.8 & 76.3 & 88.3 & 60.0 & 79.6 & 87.6 & 90.7 & 17.9 & 3.5  & 41.0 & 8.5  & 57.7 & 15.8 \\
GTE-ModernColBERT & 46.0 & 9.2  & 70.7 & 85.4 & 55.5 & 80.0 & 84.3 & 88.6 & 16.4 & 0.7  & 35.4 & 6.6  & 52.6 & 11.9 \\
jina-colbert-v2      & 11.5 & 5.7  & 3.0  & 23.4 & 18.9 & 16.0 & 13.7 & 17.2 & 0.7  & 14.4 & 6.5  & 4.0  & 26.1 & 0.6 \\
\bottomrule
\end{tabular}
\caption{Per-language MLDR results (NDCG@10) for our multilingual models and baselines. Best per column in \textbf{bold}. Language codes: ar (Arabic), en (English), fr (French), de (German), it (Italian), pt (Portuguese), es (Spanish), zh (Chinese), hi (Hindi), ja (Japanese), ko (Korean), ru (Russian), th (Thai).}
\label{tab:mldr-full}
\end{table*}

\subsection{MTEB Code}
\label{app:code}

Table~\ref{tab:code-full} reports per-task MTEB Code results for our multilingual models and other baselines. 

\begin{table*}[t]
\centering
\footnotesize
\setlength{\tabcolsep}{2pt}
\begin{tabular}{lrrrrrrrrrrrrr}
\toprule
\textbf{Model} & \textbf{Avg.} & AP & FM & FS & TC & TD & CQ & SO & SQ & CC & CN & SN & CE \\
\midrule
\multicolumn{14}{l}{\textit{Our models}} \\
\lateon     & 73.5 & 40.2 & 93.2 & 87.9 & 81.8 & 37.6 & 39.2 & 92.3 & 62.2 & \textbf{93.6} & 84.6 & 88.0 & \textbf{81.2} \\
\densemodel & 71.5 & 38.1 & 90.0 & 87.5 & 76.4 & 35.2 & 39.5 & 93.4 & 61.9 & 88.9 & 82.3 & 88.0 & 77.1 \\
\midrule
\multicolumn{14}{l}{\textit{Dense baselines}} \\
voyage-4-nano        & \textbf{76.4} & 81.0 & \textbf{94.8} & \textbf{90.0} & \textbf{90.7} & \textbf{38.7} & 36.7 & 94.3 & 62.1 & 74.9 & \textbf{85.8} & \textbf{92.4} & 75.9 \\
pplx-embed-v1-0.6b         & 75.2 & 79.0 & 85.6 & 84.7 & 86.4 & 36.8 & \textbf{43.5} & 90.7 & 58.8 & 88.1 & 82.4 & 89.8 & 76.4 \\
Qwen3-Embedding-0.6B & 73.8 & 75.9 & 90.4 & 86.5 & 86.2 & 33.0 & 41.0 & 89.3 & 59.9 & 85.1 & 84.2 & 89.9 & 63.4 \\
jina-v5-text-small   & 73.0 & 73.3 & 85.3 & 87.4 & 89.4 & 33.0 & 39.7 & 93.4 & \textbf{65.7} & 70.4 & 80.6 & 89.6 & 68.4 \\
harrier-oss-v1-0.6b  & 71.0 & 49.4 & 93.6 & 87.5 & 88.8 & 36.2 & 37.9 & \textbf{94.5} & 60.6 & 73.3 & 80.7 & 88.3 & 60.8 \\
harrier-oss-v1-270m  & 63.8 & 22.6 & 90.0 & 83.5 & 78.5 & 37.8 & 37.7 & 89.7 & 59.6 & 50.8 & 75.8 & 84.7 & 54.9 \\
jina-v5-text-nano    & 70.8 & 58.5 & 86.3 & 86.3 & 85.1 & 32.8 & 39.9 & 92.4 & 62.2 & 69.3 & 79.2 & 89.0 & 68.1 \\
embeddinggemma-300m  & 68.8 & \textbf{84.0} & 50.4 & 80.1 & 86.1 & 34.6 & 42.8 & 86.2 & 61.8 & 73.7 & 74.8 & 89.1 & 62.0 \\
granite-311m-r2      & 63.5 & 60.6 & 57.0 & 77.2 & 83.3 & 34.5 & 35.4 & 86.6 & 51.2 & 55.5 & 77.6 & 86.9 & 56.3 \\
gte-multilingual-base & 57.5 & 12.0 & 52.3 & 77.3 & 67.7 & 35.2 & 34.1 & 87.1 & 47.9 & 53.9 & 81.3 & 89.4 & 51.4 \\
arctic-embed-l-v2    & 53.2 & 9.7  & 53.2 & 74.4 & 67.6 & 29.8 & 35.6 & 86.9 & 44.6 & 48.9 & 52.4 & 77.2 & 57.8 \\
BGE-M3               & 51.5 & 14.7 & 47.9 & 69.3 & 62.6 & 29.6 & 28.7 & 80.6 & 45.3 & 53.7 & 58.3 & 77.7 & 49.5 \\
\midrule
\multicolumn{14}{l}{\textit{Late-interaction baselines}} \\
pplx-embed-v1-late-0.6b    & 63.3 & 38.3 & 70.7 & 81.4 & 82.9 & 34.5 & 38.2 & 77.8 & 55.7 & 76.4 & 44.9 & 86.2 & 72.6 \\
GTE-ModernColBERT & 54.4 & 12.1 & 56.4 & 74.9 & 75.3 & 33.1 & 32.5 & 63.7 & 55.7 & 62.7 & 40.6 & 78.5 & 67.0 \\
ColBERT-Zero         & 53.6 & 5.3  & 51.4 & 74.9 & 73.1 & 34.0 & 33.5 & 64.0 & 53.0 & 52.4 & 48.4 & 85.3 & 67.9 \\
LFM2.5-ColBERT-350M    & 51.6 & 3.6  & 51.0 & 75.8 & 73.2 & 32.3 & 30.2 & 72.1 & 53.5 & 50.9 & 37.9 & 73.4 & 65.5 \\
jina-colbert-v2      & 49.9 & 4.0  & 43.5 & 70.6 & 66.8 & 33.4 & 26.0 & 67.1 & 53.6 & 50.3 & 41.7 & 77.9 & 63.7 \\
\bottomrule
\end{tabular}
\caption{Per-task MTEB Code results (NDCG@10) for our multilingual models and baselines. Best per column in \textbf{bold}. Task acronyms: AP (AppsRetrieval), FM (CodeFeedbackMT), FS (CodeFeedbackST), TC (CodeTransOceanContest), TD (CodeTransOceanDL), CQ (CosQA), SO (StackOverflowQA), SQ (SyntheticText2SQL), CC (CodeSearchNetCCRetrieval), CN (COIRCodeSearchNetRetrieval), SN (CodeSearchNetRetrieval), CE (CodeEditSearchRetrieval).}
\label{tab:code-full}
\end{table*}

\section{Ablations}

\subsection{Knowledge Distillation}
\label{app:distillation}

We study KL-divergence distillation combined with contrastive loss, following prior work~\cite{santhanam-etal-2022-colbertv2,xiao-etal-2024-jina,takehi2025mxbaiedgecolbert}, using mxbai-rerank-large-v2 as the teacher. We vary the KL weight and the teacher and student softmax temperatures, $\tau_t$ and $\tau_s$, while fine-tuning the multi-vector model on English data. Table~\ref{tab:li-loss-ablation} reports BEIR averages. We observe that $\tau_t{=}0.1$ is optimal with a KL weight of 1. Increasing the KL weight to 10 hurts performance; varying $\tau_s$ partially recovers this loss but remains below the best configuration. Our final multi-vector recipe therefore uses $\tau_t{=}0.1$, $\tau_s{=}0.001$, and a KL weight of 1. For the single-vector model, we likewise match the student and contrastive softmax temperatures, setting $\tau_s{=}0.02$.

\begin{table*}[t]
\centering
\footnotesize
\setlength{\tabcolsep}{4pt}
\begin{tabular}{lr}
\toprule
\textbf{Objective} & \textbf{BEIR} \\
\midrule
$\mathcal{L}_{\mathrm{InfoNCE}}$ & 56.46 \\
$\mathcal{L}_{\mathrm{InfoNCE}} + \mathcal{L}_{\mathrm{KL}}$ ($\tau_t{=}0.01$, $\tau_s{=}0.001$) & 56.70 \\
$\mathcal{L}_{\mathrm{InfoNCE}} + \mathcal{L}_{\mathrm{KL}}$ ($\tau_t{=}0.1$, $\tau_s{=}0.001$) & \textbf{56.82} \\
$\mathcal{L}_{\mathrm{InfoNCE}} + \mathcal{L}_{\mathrm{KL}}$ ($\tau_t{=}1.0$, $\tau_s{=}0.001$) & 56.70 \\
$\mathcal{L}_{\mathrm{InfoNCE}} + 10\,\mathcal{L}_{\mathrm{KL}}$ ($\tau_t{=}1.0$, $\tau_s{=}0.001$) & 55.67 \\
$\mathcal{L}_{\mathrm{InfoNCE}} + 10\,\mathcal{L}_{\mathrm{KL}}$ ($\tau_t{=}1.0$, $\tau_s{=}0.01$) & 56.22 \\
$\mathcal{L}_{\mathrm{InfoNCE}} + 10\,\mathcal{L}_{\mathrm{KL}}$ ($\tau_t{=}1.0$, $\tau_s{=}0.1$) & 55.64 \\
\bottomrule
\end{tabular}
\caption{Knowledge-distillation ablation on English data and late-interaction model, reporting average BEIR \ndcg. Teacher is mxbai-rerank-large-v2. The InfoNCE softmax temperature is fixed to $0.001$.}
\label{tab:li-loss-ablation}
\end{table*}

\subsection{MeanMaxSim}
\label{app:meanmaxsim}
For multi-vector fine-tuning we adopt MeanMaxSim scoring to accommodate variable query lengths during training. We evaluate this choice comparing it to MaxSim scoring, using $\tau{=}0.02$ for MaxSim and $\tau{=}0.001$ for MeanMaxSim with InfoNCE loss to compensate for their difference in scale ($\text{MaxSim} \in [-64,64]$ vs $\text{MeanMaxSim} \in [-1,1]$). We then observe that MeanMaxSim outperforms MaxSim on BEIR ($56.46$ vs.\ $56.19$). Although the gap is small, we believe MeanMaxSim is beneficial in any case to reduce the bias around query lengths during training.

\subsection{Matryoshka Representation Learning}
\label{app:matryoshka}

\begin{figure*}[t]
\centering
\includegraphics[width=0.6\textwidth]{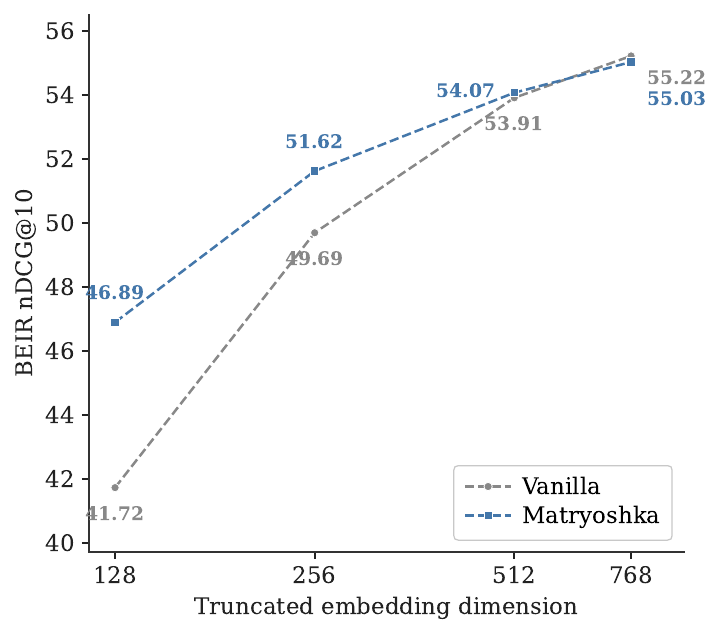}
\caption{BEIR results (average \ndcg{}) for vanilla InfoNCE and Matryoshka training across different embedding dimensions. The Matryoshka model is trained at 128, 256, 512, and 768 dimensions.}
\label{fig:matryoshka}
\end{figure*}

In order to have variable output dimensions for our dense model, we apply Matryoshka Representation Learning (MRL) on the InfoNCE loss with truncation dimensions $\{128, 256, 512, 768\}$. We compare MRL against the vanilla contrastive objective in Figure~\ref{fig:matryoshka}. MRL substantially reduces the performance drop at smaller dimensions while leaving full-dimension performance essentially unchanged, which motivates its use in our final recipe.

\subsection{Pre-Training Multilingual Mixture}
\label{app:pretraining-mixture}
To assess the impact of multilingual contrastive pre-training, we compare different  mixtures varying the proportion of English data and cross-lingual data within the multilingual portion. For each setup, we evaluate three pre-training checkpoints at $30\text{k}$, $60\text{k}$, and $90\text{k}/80\text{k}$ training steps on BEIR and on the target-language averages MIRACL\textsubscript{tgt} and MLDR\textsubscript{tgt} defined in Section~\ref{sec:multilingual-results}. We then fine-tune the checkpoints to check final performance results. Note that these early experiments were conducted before the final fine-tuning setup was selected. However, we found the results to be consistent across all setups tested.

\begin{table*}[t]
\centering
\footnotesize
\setlength{\tabcolsep}{4pt}
\begin{tabular}{l ccc ccccc}
\toprule
 & \multicolumn{3}{c}{Pre-training} & \multicolumn{5}{c}{Supervised Fine-tuning} \\
\cmidrule(lr){2-4}\cmidrule(lr){5-9}
Steps & BEIR & MIRACL\textsubscript{tgt} & MLDR\textsubscript{tgt} & BEIR & MIRACL\textsubscript{tgt} & MLDR\textsubscript{tgt} & Code & Overall \\
\midrule
\multicolumn{9}{l}{\textit{20\% English, 80\% multilingual (25\% cross-lingual)}} \\
\addlinespace[2pt]
$30\text{k}$ & $45.38$          & $44.48$ & $50.70$          & $-$ & $-$ & $-$ & $-$ & $-$ \\
$60\text{k}$ & $46.05$ & $42.79$          & $51.51$ & $55.74$ & $58.70$ & $\mathbf{67.48}$ & $\mathbf{73.24}$ & $\mathbf{63.79}$ \\
$90\text{k}$ & $45.22$          & $42.76$          & $50.70$          & $-$ & $-$ & $-$ & $-$ & $-$ \\
\midrule
\multicolumn{9}{l}{\textit{40\% English, 60\% multilingual (25\% cross-lingual)}} \\
\addlinespace[2pt]
$30\text{k}$ & $46.10$          & $45.66$ & $51.52$ & $-$ & $-$ & $-$ & $-$ & $-$ \\
$60\text{k}$ & $46.42$          & $44.10$          & $51.12$          & $55.77$ & $58.64$ & $65.75$ & $72.72$ & $63.22$ \\
$90\text{k}$ & $46.62$ & $42.80$          & $51.26$          & $-$ & $-$ & $-$ & $-$ & $-$ \\
\midrule
\multicolumn{9}{l}{\textit{60\% English, 40\% multilingual (25\% cross-lingual)}} \\
\addlinespace[2pt]
$30\text{k}$ & $47.30$ & $47.83$ & $51.48$ & $55.42$ & $\mathbf{58.86}$ & $65.91$ & $72.23$ & $63.10$ \\
$60\text{k}$ & $46.87$          & $45.19$          & $49.25$          & $-$ & $-$ & $-$ & $-$ & $-$ \\
$90\text{k}$ & $46.44$          & $42.78$          & $49.85$          & $-$ & $-$ & $-$ & $-$ & $-$ \\
\midrule
\multicolumn{9}{l}{\textit{60\% English, 40\% multilingual (60\% cross-lingual)}} \\
\addlinespace[2pt]
$30\text{k}$ & $47.65$ & $43.45$ & $49.71$ & $55.32$ & $58.64$ & $65.75$ & $72.23$ & $62.98$ \\
$60\text{k}$ & $45.59$          & $41.41$          & $49.12$          & $-$ & $-$ & $-$ & $-$ & $-$ \\
$90\text{k}$ & $47.05$          & $42.68$          & $48.66$          & $-$ & $-$ & $-$ & $-$ & $-$ \\
\midrule
\multicolumn{9}{l}{\textit{100\% English}} \\
\addlinespace[2pt]
$30\text{k}$ & $49.82$ & $34.98$ & $44.51$ & $\mathbf{55.86}$ & $58.54$ & $65.35$ & $72.16$ & $62.98$ \\
$60\text{k}$ & $48.59$          & $33.97$          & $43.34$          & $-$ & $-$ & $-$ & $-$ & $-$ \\
$80\text{k}$ & $48.51$          & $33.24$          & $44.35$          & $-$ & $-$ & $-$ & $-$ & $-$ \\
\bottomrule
\end{tabular}
\caption{Mixture ablation (\ndcg{}) on single-vector retrieval. \emph{Steps} denotes the number of pre-training steps at which the checkpoint is taken. Pre-training and Supervised Fine-tuning columns
report scores before and after supervised fine-tuning. \emph{Code} is the
MTEB Code average and \emph{Overall} the mean of the four scores for the fine-tuned models.
Best per column in \textbf{bold} for fine-tuned models.}
\label{tab:premix}
\end{table*}

For pre-training only results, increasing the proportion of English data consistently improves BEIR performance, peaking at the 100\% English mixture. MIRACL and MLDR also seem to benefit from more English data up to the 60\% English mixture, beyond which performance drops sharply.

After fine-tuning, these trends largely break down. BEIR still peaks at 100\% English, but the gap to the 20\% English mixture is minimal, showing that pre-training performance is not a reliable indicator of final performance. MIRACL remains consistent with pre-training, peaking at 60\% English, which confirms, somewhat unexpectedly, that more English data helps MIRACL, but the gap to the other mixtures is again small. MLDR instead peaks at 20\% English with a considerable gap over the 60\% English mixture, reversing the pre-training picture. Increasing the proportion of cross-lingual data does not significantly affect final results. Overall, the 20\% English mixture achieves the best overall result after fine-tuning, although the margins between mixtures are small. We adopt the 20\% English mixture for our final recipe.

Since the $20\%$ English, $25\%$ cross-lingual mixture yields the best final performance for the dense model, we adopt it to examine the effect of the number of pre-training steps on the late-interaction model. We fine-tune three pre-training checkpoints, taken at $40\text{k}$, $60\text{k}$, and $90\text{k}$ steps, using our late-interaction supervised fine-tuning recipe. Ablation results are shown in Table~\ref{tab:li-ckpt}. Final performance is largely insensitive to the number of pre-training steps, with only small differences across checkpoints and the $40\text{k}$ checkpoint attains the best overall score. We thus select it as our pre-training checkpoint for the final late-interaction recipe.

\begin{table*}[t]
\centering
\footnotesize
\setlength{\tabcolsep}{6pt}
\begin{tabular}{l ccccc}
\toprule
Steps & BEIR & MIRACL\textsubscript{tgt} & MLDR\textsubscript{tgt} & Code & Overall \\
\midrule
$40\text{k}$ & $\mathbf{57.56}$ & $\mathbf{65.61}$ & $87.69$          & $73.48$          & $\mathbf{71.09}$ \\
$60\text{k}$ & $57.42$          & $65.47$          & $\mathbf{87.81}$ & $73.55$          & $71.06$ \\
$90\text{k}$ & $57.26$          & $65.24$          & $87.57$          & $\mathbf{73.59}$ & $70.92$ \\
\bottomrule
\end{tabular}
\caption{Pre-training checkpoint ablation (\ndcg{}) on late-interaction retrieval, using the final $20\%$ English, $25\%$ cross-lingual mixture. \emph{Steps} denotes the number of pre-training steps at which the checkpoint is taken, before fine-tuning with our final recipe. \emph{Code} is the MTEB Code average and \emph{Overall} the mean of the four scores. Best per column in \textbf{bold}.}
\label{tab:li-ckpt}
\end{table*}

\section{Training Hyperparameters}
\label{app:hyperparams}
Table~\ref{tab:app-hyperparams} summarizes the hyperparameters of the contrastive pre-training setup, and the supervised fine-tuning and distillation settings of the dense and late-interaction models.

\begin{table*}[h]
\centering
\small
\begin{tabular}{ll}
\toprule
\multicolumn{2}{l}{\textit{Contrastive Pre-Training}} \\
\addlinespace[2pt]
Initialization       & \baseencoder \\
Objective            & InfoNCE \\
Negatives            & In-batch negatives \\
Scoring              & Dense: CLS pooling + cosine; LI: MaxSim \\
Max context length       & Dense: $1{,}024$ tokens; LI: $32$ query / $300$ document tokens \\
Temperature          & $0.02$ \\
Optimizer            & AdamW \\
LR                   & $8\!\times\!10^{-5}$ \\
Epochs               & $1$ \\
Global batch size    & $16{,}416$ \\
Batch sampler        & Multinomial, $\alpha{=}0.5$ \\
Training mixed precision & bfloat16 (Flash-attention 2 enabled) \\
Hardware             & $32\!\times\!$H100 80\,GB \\
\midrule
\multicolumn{2}{l}{\textit{Supervised Fine-Tuning}} \\
\addlinespace[2pt]
Initialization       & 20\% English, 80\% multilingual (25\% X-lingual) \\
Pre-training steps   & Dense: $60$k; LI: $40$k \\
Scoring              & Dense: CLS pooling + cosine; LI: MeanMaxSim, no query expansion \\
Embedding dimension  & Dense: $768$; LI: $128$ (per token) \\
Prefixes             & Dense: \texttt{"query: "} / \texttt{"document: "}; LI: \texttt{"[Q] "} / \texttt{"[D] "} \\
Loss                 & Dense: Matryoshka InfoNCE + KL-div; LI: InfoNCE + KL-div \\
Negatives            & 7 sampled out of $10$ mined per query (NV threshold $0.95$) \\
Contrastive temperature & Dense: $0.02$; LI: $0.001$ \\
KL temperatures      & teacher $0.1$; student: Dense $0.02$, LI $0.001$ \\
Matryoshka           & Dense: dims $\{768, 512, 256, 128\}$; LI: not applied \\
KD teacher           & mxbai-rerank-large-v2 \\
Optimizer            & AdamW \\
LR                   & $3\!\times\!10^{-6}$ \\
Epochs               & $1$ \\
Global batch size    & $128$ \\
Max context length   & $8{,}192$ tokens (queries and documents) \\
Batch sampler        & multinomial, $\alpha{=}0.5$ \\
Training mixed precision & bfloat16 (Flash-attention 2 enabled) \\
Hardware             & $8\!\times\!$H100 80\,GB \\
\bottomrule
\end{tabular}
\caption{Training hyperparameters for contrastive pre-training and supervised fine-tuning (SFT) of the dense and late-interaction (LI) models.}
\label{tab:app-hyperparams}
\end{table*}

\section{Evaluation Hyperparameters}
\label{app:eval-params}

Table~\ref{tab:eval-tasks} lists the exact MTEB tasks used in our evaluation,
together with the evaluation settings. Our models and all late-interaction
baselines are evaluated in bfloat16, enabling FlashAttention-2, with a maximum
sequence length of 8{,}192 tokens for both queries and documents, while dense
baselines are evaluated with their original precision and context length. Our dense models are evaluated with the ``query: '' and ``document: '' prefixes and our late-interaction
models with the ``[Q] '' and ``[D] '' prefixes; all baselines keep their
original prompts and, where applicable, adapters.

\begin{table*}[t]
\centering
\small
\begin{tabular}{lp{0.55\linewidth}}
\toprule
\textbf{BEIR} \\
\midrule
MSMARCO \\
FEVER \\
ClimateFEVER \\
HotpotQA \\
DBPedia \\
NQ \\
QuoraRetrieval \\
CQADupstackRetrieval \\
Touche2020 \\
TRECCOVID \\
FiQA2018 \\
SCIDOCS \\
ArguAna \\
SciFact \\
NFCorpus \\
\midrule
\textbf{Multilingual} \\
\midrule
MIRACLRetrievalHardNegatives \\
MultiLongDocRetrieval \\
\midrule
\textbf{Code} \\
\midrule
SyntheticText2SQL \\
CodeFeedbackMT \\
CodeFeedbackST \\
COIRCodeSearchNetRetrieval \\
CodeSearchNetCCRetrieval \\
CodeEditSearchRetrieval \\
StackOverflowQA \\
CosQA \\
AppsRetrieval \\
CodeSearchNetRetrieval \\
CodeTransOceanContest \\
CodeTransOceanDL \\
\midrule
\textbf{Evaluation settings} \\
\midrule
Framework versions          & MTEB 2.12.25 (LI) / 2.10.12 (Dense), PyLate 1.5.0, FastPlaid 1.4.7, Sentence Transformers 5.4.1 \\
Our models          & bfloat16, 8{,}192-token max sequence length (queries and documents) \\
LI baselines        & bfloat16, 8{,}192-token max sequence length (queries and documents) \\
Dense baselines     & original precision and context length \\
Prefixes and adapters           & ``query: '' / ``document: '' (our Dense), ``[Q] '' / ``[D] '' (our LI); baselines keep their original prompts and adapters \\
Fast Plaid search parameters  & nbits=4, n\_ivf\_probe=8, n\_full\_scores=8{,}192 \\
\bottomrule
\end{tabular}
\caption{Retrieval tasks used in our evaluation, grouped by benchmark, with
the evaluation settings. Task names are the exact MTEB identifiers; documents
longer than the sequence-length limit are truncated.}
\label{tab:eval-tasks}
\end{table*}

\end{document}